\documentclass[11pt]{article}

\usepackage[final]{acl}

\usepackage{times}
\usepackage{latexsym}

\usepackage[T1]{fontenc}

\usepackage[utf8]{inputenc}

\usepackage{microtype}

\usepackage{inconsolata}

\usepackage{graphicx}
\usepackage{subcaption}
\usepackage{algorithm}
\usepackage{amsmath}
\usepackage{booktabs} 
\usepackage{longtable}
\usepackage[normalem]{ulem}
\usepackage{hyperref}
\usepackage{arydshln}
\usepackage{xcolor}

\newcommand{\rp}[1]{\textcolor{blue}{#1}}
\newcommand{\mee}[1]{\textcolor{red}{#1}}

\usepackage{comment}

\usepackage{amssymb}

\title{Cross-Lingual Activation Steering for Multilingual Language Models}

\author{
Rhitabrat Pokharel\textsuperscript{1},
Ameeta Agrawal\textsuperscript{1},
Tanay Nagar\textsuperscript{2}\\[1ex]
\textsuperscript{1}Department of Computer Science, Portland State University, USA\\
\texttt{\{pokharel,ameeta\}@pdx.edu}\\[0.5ex]
\textsuperscript{2}Carnegie Mellon University\\
\texttt{tnagar@andrew.cmu.edu}
}

\begin{document}
\maketitle
\begin{abstract}
Large language models exhibit strong multilingual capabilities, yet significant performance gaps persist between dominant and non-dominant languages. Prior work attributes this gap to imbalances between shared and language-specific neurons in multilingual representations. We propose Cross-Lingual Activation Steering (CLAS), a training-free inference-time intervention that selectively modulates neuron activations. We evaluate CLAS on several multilingual downstream tasks   and achieve improvements while maintaining high-resource language performance. We discover that effective transfer operates through functional divergence rather than strict alignment, and performance gains correlate with increased language cluster separation. Our results demonstrate that targeted activation steering can unlock latent multilingual capacity in existing models without modification to model weights.
\end{abstract}

\section{Introduction}

Large Language Models (LLMs) perform well on many multilingual tasks, but a substantial performance gap remains between dominant (high-resource) and non-dominant (low-resource) languages. This gap is largely attributed to the heavy skew of pre-training corpora toward dominant languages, which enables models to develop richer representations in those languages \citep{zhang-etal-2024-enhancing-multilingual, xu2025survey}. 

To address this problem, prior work has explored data- and training-based solutions such as multilingual instruction tuning \cite{chen-etal-2024-monolingual, shaham-etal-2024-multilingual}, supervised fine-tuning \cite{chen-etal-2024-breaking}, and model alignment using smaller multilingual datasets \cite{she-etal-2024-mapo, gao-etal-2024-multilingual, pokharel2025capo}. Although effective, these methods depend on annotated datasets or additional training. A complementary line of work studies multilingual behavior at the neuron level as a lightweight alternative. Researchers have identified shared, partially shared, and language-specific neurons \citep{wang2024sharing,tang-etal-2024-language}, and shown that shared neurons and overlapping subspaces in middle and upper layers play a central role in cross-lingual transfer \citep{tezuka-inoue-2025-transfer,xu-etal-2025-linguistic}. At the same time, direct manipulation of language-specific neurons has produced mixed results \citep{mondal-etal-2025-language}. Other work suggests that multilingual models often reason through an English-like latent space before generating outputs in the target language \citep{etxaniz-etal-2024-multilingual,zhao2024how}.


\begin{figure*}[!t]
    \centering
    \includegraphics[width=1.0\textwidth]{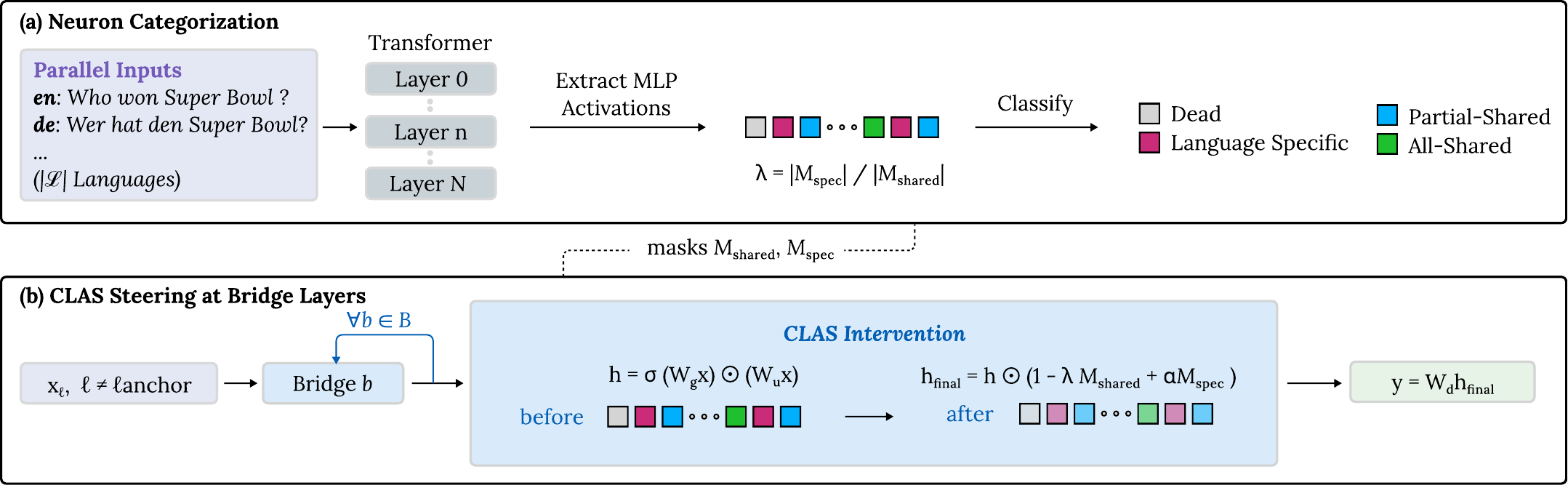}
    \caption{Overview of CLAS, with two high level steps (1) Neuron Categorization and (2) Activation Steering.} \vspace{-0.5cm}
    \label{fig:overall_working_clas}
\end{figure*}

Building on these insights, we introduce Cross-Lingual Activation Steering (CLAS), a test-time neuron-level intervention that rebalances shared and language-specific representations during inference. CLAS gently suppresses the dominant shared neurons that otherwise overwhelm finer linguistic signals, while selectively amplifying or attenuating language-specific neurons to steer the model toward the desired target language. This targeted reweighting improves performance on non-dominant languages without any additional data or parameter updates. Figure~\ref{fig:overall_working_clas} provides an overview of our proposed approach.

{While \citet{mondal-etal-2025-language} also study test-time neuron interventions, their method has an important limitation: it replaces selected neuron values with fixed constants, ignoring the input and removing useful variation. In contrast, our approach keeps the model’s original signal and adjusts it proportionally, so the result still depends on the input. Similarly, INCLINE \cite{wang-etal-2025-bridging} requires tuning a separate hyperparameter for each language and task, which is time- and resource-intensive, while our method only needs tuning once per task and modifies a small number of layers.}

Our main contributions are as follows:
\begin{itemize}
\itemsep0em 
    \item We propose CLAS, a training-free activation steering mechanism that selectively modulates neurons to enhance cross-lingual transfer during inference\footnote{The code will be made publicly available at \url{https://github.com/PortNLP/CLAS}.}.
    \item We demonstrate CLAS effectiveness across diverse tasks and models, achieving improvements on both classification and generation benchmarks while maintaining anchor language stability.
     \item We conduct comprehensive analysis of cross-lingual representations and find that effective transfer is driven by functional divergence rather than proximity to the anchor language.
\end{itemize}
\section{Cross-Lingual Activation Steering (CLAS)}
CLAS consists of three stages: (i) construct parallel inputs across languages, (ii) summarize neuron behavior using simple activation statistics and group neurons into coarse categories, and (iii) apply a lightweight steering rule that modulates activations at inference time. 

\subsection{Preliminaries}
We define a configuration $\mathcal{C} = \{\mathcal{L},\, \ell_{\text{anchor}},\, \mathcal{B},\,\, \alpha \}$, where $\mathcal{L}$ is the set of languages used for analysis, $\ell_{\text{anchor}}$ is an anchor language (typically the model's strongest language),  $\mathcal{B}$ is the set of ``bridge layers'' selected for intervention, and $\alpha$ is the hyperparameter that controls the strength of boosting or suppressing selected activations during steering.

\subsection{Parallel Input Construction}
To analyze neuron behavior across languages, we require aligned multilingual inputs that express the same underlying content. For each sample index $i$, we construct a set of parallel inputs $
\mathbf{x}^{(i)} = \{x^{(i)}_{\ell} \mid \ell \in \mathcal{L}\}$, 
where each $x^{(i)}_{\ell}$ is the same text expressed in language $\ell$. These parallel inputs are used only for measuring neuron activations and do not update the model in any way.

\subsection{Neuron Statistics and Categorization}\label{sec:neuron}

\begin{figure}[!t]
    \centering
    \vspace{-0.3cm}
    \begin{subfigure}{0.52\textwidth}
        \centering
        \includegraphics[width=\textwidth]{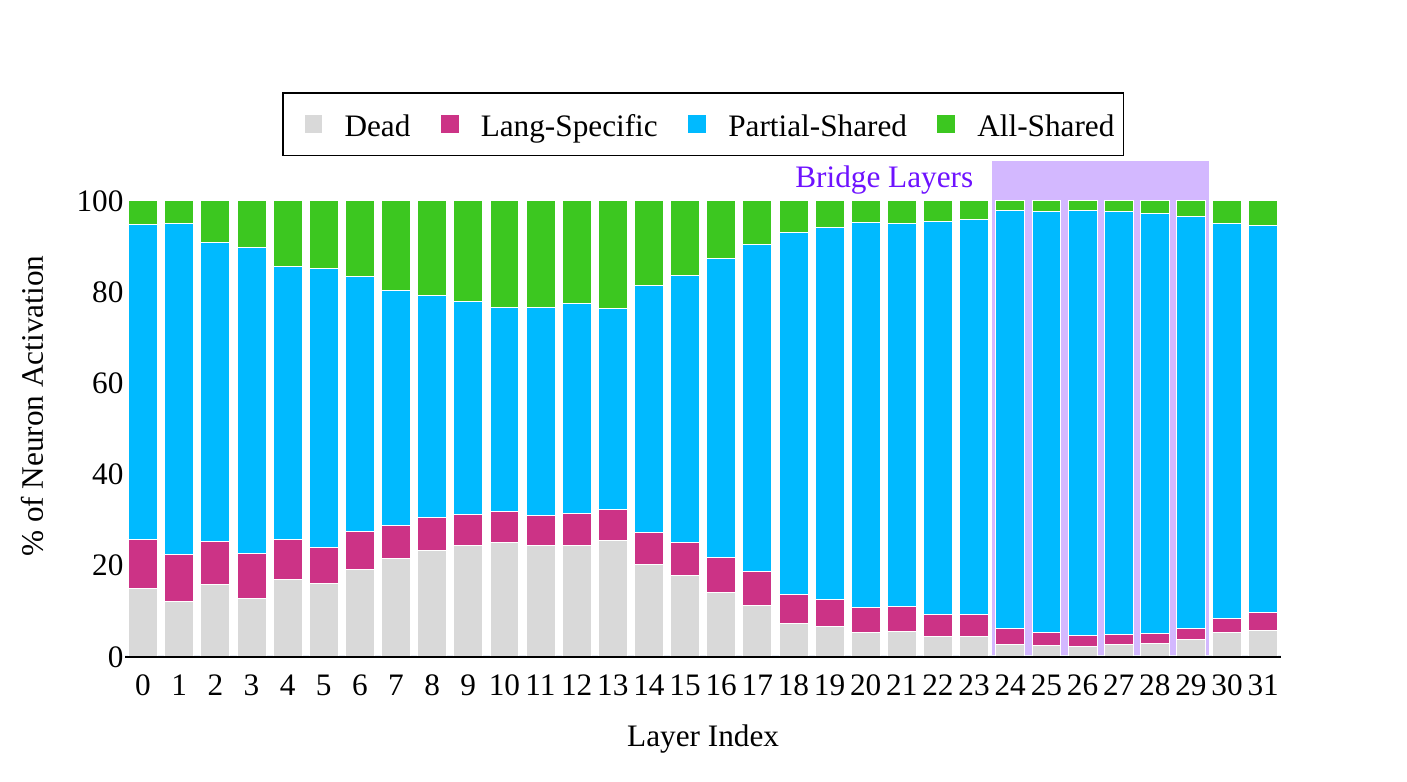}
        \caption{Llama}
        \label{fig:llama_xquad}
    \end{subfigure}\vspace{-0.1cm}
    \begin{subfigure}{0.52\textwidth}
        \centering
        \includegraphics[width=\textwidth]{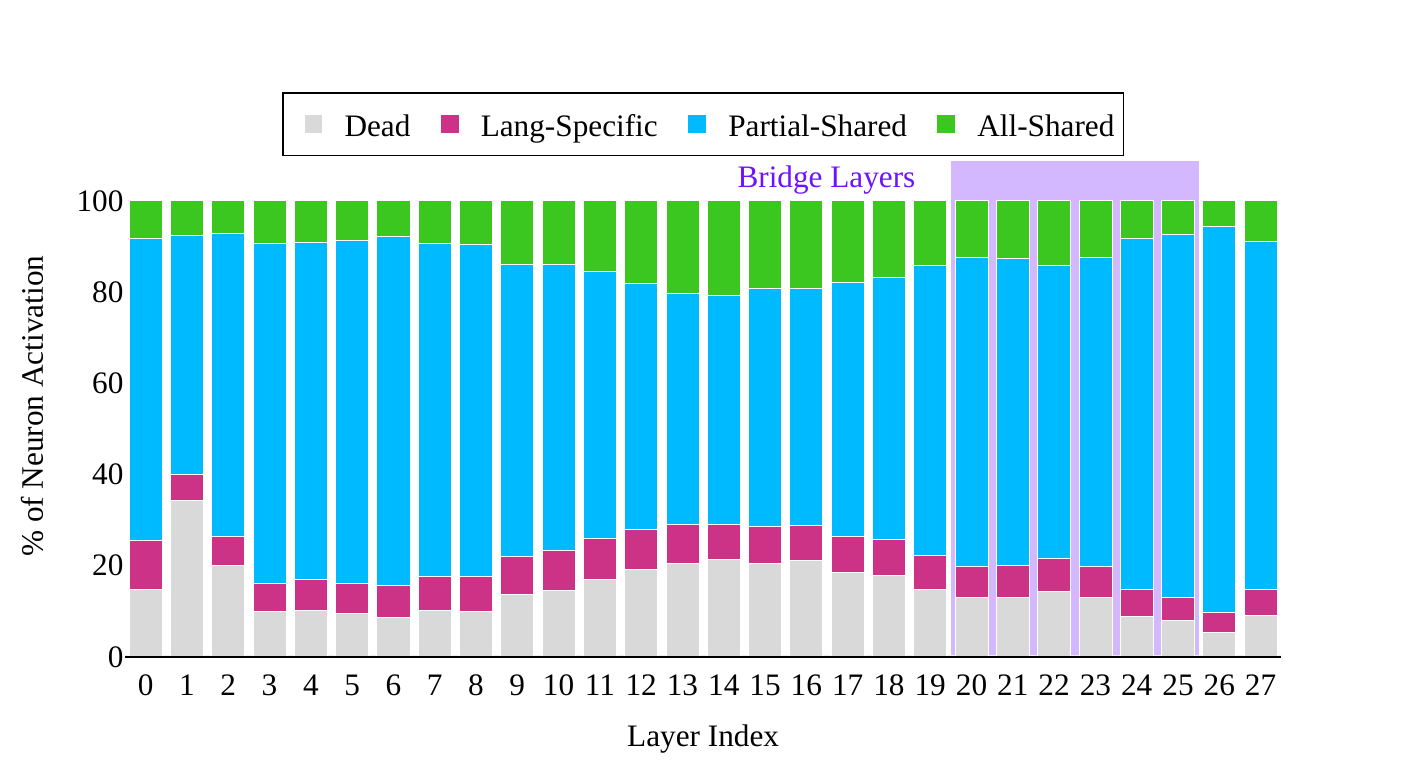}
        \caption{Qwen}
        \label{fig:qwen_xquad}
    \end{subfigure}

    \caption{Distribution of types of neuron per layer across models. Llama has a total of 32 layers and Qwen has 28 layers. The bridge layers (purple shade) are selected near the final layers where the partial-shared language activations are the highest.}\vspace{-0.5cm}
    \label{fig:neuron_activity}
\end{figure}

To understand the model's multilingual structure and identify where CLAS should intervene, we first analyze neuron activations across languages and layers. Following  \cite{wang2024sharing}, we group neurons into four mutually exclusive categories: dead (never active), language-specific (active for one language), partial-shared (active for some languages), and all-shared (active for all languages). However, we differ from prior work in how these categories are computed. \citet{wang2024sharing} assign categories at the instance level: a neuron is considered all-shared if it activates for all languages on a single parallel example. As a result, neuron categories can vary across inputs and tasks. In contrast, we adopt a dataset-level view. We compute the mean activation of each neuron using 100 parallel samples  from the XQuAD dataset across 12 languages and two models (Llama 3.1 8B and Qwen 2.5 7B) and use this aggregate statistic to assign each neuron to a single category. This makes each neuron’s category a stable property of the model and language set rather than something that changes across examples. {We observe a similar phenomenon with XNLI dataset.} 

\paragraph{Layer-wise Activation Patterns}
For each layer, we compute the percentage of neurons in each category. Figure~\ref{fig:neuron_activity} summarizes the resulting distributions. We observe several consistent patterns:



\noindent \textbf{\em Early layers.} Early layers contain a high proportion of partial-shared neurons, suggesting that these layers are involved in mapping language-specific or language-related features into shared representations \citep{tang-etal-2024-language, zhang2025does}. 

\noindent \textbf{\em Middle layers.} In the middle layers, we observe an increase in all-shared neurons alongside a rise in dead neurons and a reduction in partial-shared neurons. This is consistent with prior findings that mid-level representations become more language-agnostic and semantically focused \citep{zhang2025does, xu-etal-2025-linguistic}.
    
\noindent  \textbf{\em Final layers.} In the upper layers, partial-shared neurons become more prevalent again, and dead neurons decrease, suggesting that language-specific processing re-emerges near the output as the model prepares to generate text in the target language \citep{tang-etal-2024-language, zhang2025does}.

These observations are descriptive and do not imply that these categories are sharply separated or functionally pure; rather, they provide a coarse summary of how multilingual structure is distributed across layers.

\paragraph{Selecting Bridge Layers}

Based on these statistics, we select a small set of \emph{bridge layers} for intervention. In both LLaMA and Qwen, layers 24-29 and 20-25, respectively, exhibit relatively high proportions of partial-shared neurons together with relatively low proportions of dead and language-specific neurons. We therefore hypothesize that intervening in these layers provides a useful balance: representations are shared enough to support cross-lingual steering, while not yet so specialized that small perturbations directly disrupt surface generation. We exclude the final two layers from intervention, as these layers are known to be strongly specialized for producing language-specific outputs and are particularly sensitive to perturbation \citep{zhao2024large}. All layer selections are fixed prior to evaluation and are not tuned on test data.






\subsection{Activation Steering via CLAS} \label{sec:steering}

Given an input $x_{\ell}$ in language $\ell$, the model computes an intermediate MLP
activation using the standard SwiGLU transformation:
\[
\mathbf{h} = \sigma(W_g \mathbf{x}) \odot (W_u \mathbf{x}),
\]
where $W_g$ and $W_u$ are the gating and up-projection matrices, and $\sigma$ is the
nonlinear activation function. 

For non-anchor languages $\ell \neq \ell_{\text{anchor}}$, CLAS applies a lightweight, deterministic modification to this activation using masks derived from neuron categories to adjust the relative contribution of shared and language-specific neurons.

\paragraph{Adaptive Masking and Heuristic}
To effectively steer the model's intermediate representations, we must account for the structural imbalance between neuron categories. Let $M_{\text{shared}}$ denote the combined mask for all shared neurons (i.e., the union of partial and all-shared neurons), and let $M_{\text{spec}}$ denote the mask for language-specific neurons. In the target layers (e.g., the bottom layers), shared neurons inherently dominate the representation space ($|M_{\text{shared}}| \gg |M_{\text{spec}}|$). 

To balance our intervention and prevent the abundant shared neurons from overpowering the scarce language-specific signals, we introduce an adaptive scaling factor $\lambda$, defined as the ratio of active specific neurons to shared neurons:
\[
\lambda = \frac{|M_{\text{spec}}|}{|M_{\text{shared}}|},
\]
where $|\cdot|$ represents the count of active neurons within a given mask. Because shared neurons form the majority of the active dimensions, $\lambda$ strictly remains a fraction ($\lambda \ll 1$). This formulation ensures that the intervention strength is not static, but naturally grounded in the precise architectural makeup of each individual layer.

\paragraph{Inference-Time Steering Intervention}
We apply a single-step multiplicative intervention designed to reweight the neuron contributions:
\[
\mathbf{h}_{\text{final}} = \mathbf{h} \odot \left( \mathbf{1} - \lambda M_{\text{shared}} + \alpha M_{\text{spec}} \right),
\]
where $\alpha$ is a configurable hyperparameter that dictates the amplification or suppression of language-specific neurons.

Subtracting $\lambda M_{\text{shared}}$ gently suppresses the shared neurons, preventing them from drowning out finer language-specific signals. Adding $\alpha M_{\text{spec}}$ then explicitly amplifies or suppresses the language-specific neurons depending on the desired steering direction: a positive $\alpha$ pushes the representation toward the target language, while a negative $\alpha$ pulls it away. The adjusted activation is then passed through the down-projection:
\[
\mathbf{y} = W_d \mathbf{h}_{\text{final}}.
\]


\subsection{Anchor Language Handling}

For the anchor language $\ell_{\text{anchor}}$ (i.e. English), no modification is applied: $\mathbf{h}_{\text{final}} = \mathbf{h}$. We keep the anchor activations untouched because the anchor language serves as a stable reference point for cross-lingual alignment. Modifying it would risk introducing unnecessary distortion into a representation that is already well supported by the model. All adjustments are therefore applied only to non-anchor languages, allowing them to shift relative to a fixed reference.

\section{Experimental Setting}
{To assess CLAS's performance, we detail our setup using 2 models and 5 datasets.}

\subsection{Models}
We use LLaMA 3.1 8B Instruct \cite{grattafiori2024llama} and Qwen 2.5 7B Instruct \cite{qwen25}. Both models are kept frozen; CLAS is applied only at inference time and does not modify model parameters. 

\subsection{Datasets and Languages}
We evaluate on the following five benchmarks, covering a total of 21 languages. More details are presented in Appendix~\ref{app:dataset}.

\begin{table*}[!t]
\centering
\setlength{\tabcolsep}{8pt}
\small
\begin{tabular}{lrrrr|rrrrr}
\toprule
&\multicolumn{4}{c}{Llama} &\multicolumn{4}{c}{Qwen} \\
\cmidrule{2-9}
&\textbf{Base} &\textbf{Int $\mu$} &\textbf{INCLINE} &\textbf{CLAS (Ours)} &\textbf{Base} &\textbf{Int $\mu$} &\textbf{INCLINE} &\textbf{CLAS (Ours)} \\
\midrule
XNLI &42.42 &42.19 &39.19 &\textbf{45.03} (+2.61) &59.08 &59.05 &41.34 &\textbf{59.29} (+0.21) \\
XStoryCloze &83.38 &82.97 &79.05 &\textbf{85.58} (+2.20) &78.68 &78.96 &61.08 &\textbf{79.70} (+1.02) \\
PAWS-X &65.70 &62.87 &59.58 &\textbf{66.00} (+0.30) &78.58 &79.03 &68.30 &\textbf{79.07} (+0.49) \\
HarryPotter &51.59 &47.84 &50.75 &\textbf{52.59} (+1.00) &53.67 &49.58 &28.42 &\textbf{55.50} (+1.83) \\
XQuAD &26.85 &26.98 &\textbf{33.50} &32.10 (+5.25)	&32.73	&21.78	&18.38	&\textbf{33.94} (+1.21)\\
\bottomrule
\end{tabular}
\caption{The performance on different benchmarks across Llama and Qwen. The best performing model is in \textbf{bold}. The improvements of CLAS over base are shown within parentheses.} \vspace{-0.3cm}
\label{tab:classification_results}
\end{table*}

\noindent {\bf XNLI} \cite{conneau2018xnli} is a multilingual natural language inference dataset in 15 languages. We prompt the model with a premise and hypothesis, reporting accuracy based on the highest-probability target token:  Entailment,  Neutral, or Contradiction. 

\noindent \textbf{XStoryCloze} \cite{lin2022few} is a multilingual commonsense reasoning dataset based on StoryCloze. Given a four-sentence context and two candidate endings, the model selects the correct ending. We evaluate on 11 languages and report accuracy. 

\noindent \textbf{PAWS-X} \cite{yang2019pawsxcrosslingualadversarialdataset} is a cross-lingual paraphrase identification dataset with adversarial sentence pairs that have high lexical overlap but different meanings. We evaluate on 4 languages and report accuracy.

\noindent \textbf{HarryPotter Quiz}  \cite{chua2025crosslingualcapabilitiesknowledgebarriers} is a cross-lingual factual knowledge benchmark that tests whether models can access the same knowledge across languages. We evaluate on 5 languages and report accuracy.

\noindent {\bf XQuAD} \cite{artetxe2019xquad} is a multilingual question answering dataset in 12 languages. Each instance requires the model to predict the correct answer span within the provided text. The results are reported using the token-level F1 score.


\subsection{Baselines}
We consider three baselines: \textbf{(1)} unmodified base model, \textbf{(2)} Int~$\mu$ \citep{mondal-etal-2025-language}, which substitutes the model's intermediate representations with mean activations during intervention; and \textbf{(3)} INCLINE \citep{wang-etal-2025-bridging} which learns alignment matrices applied during inference to project the intermediate representations of low-performing languages toward the representation space of a high-performing anchor language. {INCLINE uses language-specific hyperparameter and fully reproducing this setup would require tuning a separate parameter for each language, which is beyond the scope of our experiments. Instead, we fix $\alpha=1$ across all settings as this was reported to perform well for most languages \cite{wang-etal-2025-bridging}}.



\section{Results and Discussion}
{In this section, we discuss the results of the experiments with CLAS and how it improves multilingual performance.}


\begin{table*}[!t]
\centering
\small
\begin{tabular}{lrrrrrrrrrrrr}
\toprule
& en & ar & es & eu & hi & id & my & ru & sw & te & zh & \textbf{Avg.} \\
\midrule
\multicolumn{13}{l}{\textit{Llama}} \\
\cmidrule(lr){1-13}
Base   &92.26 &88.09 &91.07 &70.02 &90.07 &87.49 &74.26 &91.40 &71.01 &79.68 &90.73 &83.38 \\
Int $\mu$  &92.39 &89.01 &91.33 &71.74 &86.30 &84.98 &74.39 &91.79 &71.08 &78.89 &90.14 &82.97 \\
INCLINE &92.39 &90.21 &95.10 &67.77 &84.58 &88.88 &52.75 &92.52 &71.14 &52.81 &94.77 &79.05 \\
CLAS    &92.26 &85.44 &\textbf{94.77} &\textbf{74.92} &89.61 &\textbf{92.12} &73.13 &\textbf{94.04} &\textbf{78.09} &78.95 &\textbf{94.71} &\textbf{85.58} \\
\addlinespace
\multicolumn{13}{l}{\textit{Qwen}} \\
\cmidrule(lr){1-13}
Base    &97.68 &91.46 &95.10 &52.95 &87.03 &93.58 &60.69 &94.97 &54.73 &61.02 &95.30 &78.68 \\
Int $\mu$  &97.68 &91.59 &95.10 &52.42 &87.09 &93.91 &61.68 &94.37 &55.79 &62.41 &95.23 &78.96 \\
INCLINE &97.68 &90.54 &52.88 &52.68 &51.62 &53.67 &52.81 &93.71 &52.81 &57.31 &52.81 &61.08 \\
CLAS    &97.68 &\textbf{91.53} &\textbf{95.96} &\textbf{53.34} &\textbf{87.62} &\textbf{93.85} &\textbf{61.68} &\textbf{95.70} &\textbf{56.52} &\textbf{65.32} &\textbf{95.50} &\textbf{79.70} \\
\bottomrule
\end{tabular}
\caption{Performance on XStoryCloze (accuracy) across multiple languages and models. For each language, \textbf{bold} represents the cases where CLAS beats the base model.}\vspace{-0.3cm}
\label{tab:xstory_results}
\end{table*}

\begin{table*}[!t]
\centering
\setlength{\tabcolsep}{5pt}
\small
\begin{tabular}{lrrrrrrrrrrrrr}
\toprule
 & en & ar & de & el & es & hi & ro & ru & th & tr & vi & zh & \textbf{Avg.} \\
\midrule
\multicolumn{14}{l}{\textit{Llama}} \\
\cmidrule(lr){1-14}
Base   &23.47 &23.20 &38.54 &25.18 &34.23 &29.47 &30.01 &27.32 &19.23 &23.74 &34.91 &12.84 &26.85 \\
Int $\mu$  &23.47 &21.25 &33.08 &26.68 &35.31 &29.39 &36.98 &27.25 &15.83 &21.64 &40.00 &12.90 &26.98 \\
INCLINE &34.96 &20.39 &45.79 &35.68 &44.88 &34.02 &34.13 &21.87 &26.87 &44.18 &36.01 &23.19 &33.50 \\
CLAS    &23.47 &\textbf{27.03} &\textbf{50.86} &23.71 &\textbf{50.57} &28.32 &\textbf{41.84} &\textbf{34.87} &\textbf{21.90} &\textbf{30.38} &\textbf{37.49} &\textbf{14.77} &\textbf{32.10} \\
\addlinespace
\multicolumn{14}{l}{\textit{Qwen}} \\
\cmidrule(lr){1-14}
Base    &48.88 &28.73 &34.77 &37.06 &38.53 &33.12 &34.48 &27.58 &28.51 &29.19 &31.67 &20.24 &32.73 \\
Int $\mu$  &14.53 &23.11 &11.99 &48.95 &35.43 &17.83 &31.31 &17.39 &12.63 &21.25 &15.75 &11.20 &21.78 \\
INCLINE &29.91 &25.10 &3.40 &7.61 &13.37 &0.14 &20.87 &22.47 &39.56 &20.47 &33.94 &3.66 &18.38 \\
CLAS    &48.88 &28.29 &36.03 &\textbf{38.91} &\textbf{40.53} &25.98 &\textbf{38.13} &\textbf{27.93} &\textbf{29.49} &\textbf{30.28} &\textbf{35.86} &\textbf{26.92} &\textbf{33.94} \\
\bottomrule
\end{tabular}
\caption{The performance (F1 scores) on the XQuAD dataset across multiple languages and models. For each language, \textbf{bold} represents the cases where CLAS beats the base model.} \vspace{-0.5cm}
\label{tab:xquad_results}
\end{table*}

\subsection{Main Results}
Table~\ref{tab:classification_results} summarizes the results on five benchmarks, comparing CLAS against the baselines. Across all the benchmarks, CLAS consistently improves accuracy over the baselines. For Llama, CLAS yields substantial gains on XNLI (+2.61) and XStoryCloze (+2.20), while moderate enhancements on HarryPotter (+1.00) and PAWS-X (+0.30). For Qwen, the improvements are generally smaller but remain  positive across all tasks, peaking on HarryPotter (+1.83) and XQuAD (+1.21). This narrower margin of improvement reflects Qwen’s inherently stronger baseline multilingual capabilities, where CLAS acts more conservatively to provide stable, consistent enhancements without disrupting existing knowledge. Crucially, CLAS demonstrates significantly greater reliability than existing intervention baselines. While Int~$\mu$ exhibits mixed results generally degrading Llama's performance and yielding only marginal or negative shifts on Qwen, INCLINE suffers from performance degradation across both models. This regression is particularly drastic on Qwen, where INCLINE causes accuracy to drop steeply (e.g., from 53.67 to 28.42 on HarryPotter). In contrast, CLAS achieves positive gains across all tasks and both model families.

\subsection{Language-Level Performance}
To better understand where the overall gains come from, we analyze performance by language on XStoryCloze (Table~\ref{tab:xstory_results}) and XQuAD (Table~\ref{tab:xquad_results}). As intended by design, CLAS preserves the anchor language  English performance which remains essentially unchanged across both models and tasks.

On XStoryCloze (discriminative multiple-choice), CLAS improves many non-English languages. For Llama, gains are large and broadly distributed, e.g., Swahili (+7.08), Basque (+4.90), Indonesian (+4.63), and Chinese (+3.98), with small regressions in a few cases (e.g., Arabic and Burmese). For the stronger Qwen model, CLAS acts as a consistent refinement, yielding targeted improvements (notably Telugu +4.30 and Swahili +1.79) without regressions. This stability contrasts with INCLINE, which collapses on Qwen for XStoryCloze, dropping performance to roughly the low-50s across many languages. Results for the remaining datasets are in Appendix~\ref{sec:rem_lang_peroformance}.

On XQuAD (generative QA), effects are more uneven across languages, consistent with generation being more sensitive to representation shifts. CLAS again preserves the anchor, while INCLINE is inconsistent, boosting English on Llama (23.47 $\rightarrow$ 34.96) but degrading it on Qwen (48.88 $\rightarrow$ 29.91). For Llama, CLAS delivers a strong +5.25 average F1 gain, driven by large improvements in Spanish (+16.34), German (+12.32), and Romanian (+11.83). For Qwen, CLAS gives a modest +1.21 average F1 improvement, combining clear wins in some languages (e.g., Chinese and Vietnamese) with regressions in others (notably Hindi), highlighting that test-time interventions are more stable on discriminative tasks than on generation.





\begin{figure}[!t]
    \centering

    \begin{subfigure}{0.5\textwidth}
        \centering
        \includegraphics[width=0.7\textwidth]{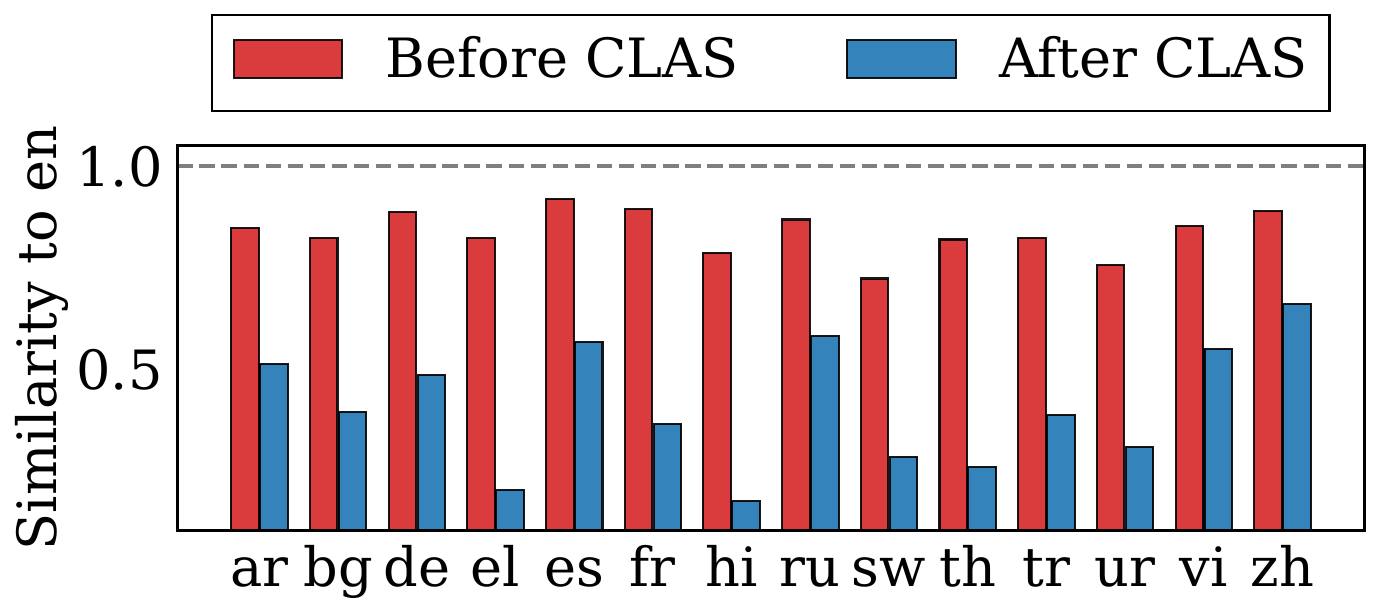}
        \caption{XNLI}
        \label{fig:analysis_xnli_llama}
    \end{subfigure}

    \begin{subfigure}{0.5\textwidth}
        \centering
        \includegraphics[width=0.7\textwidth]{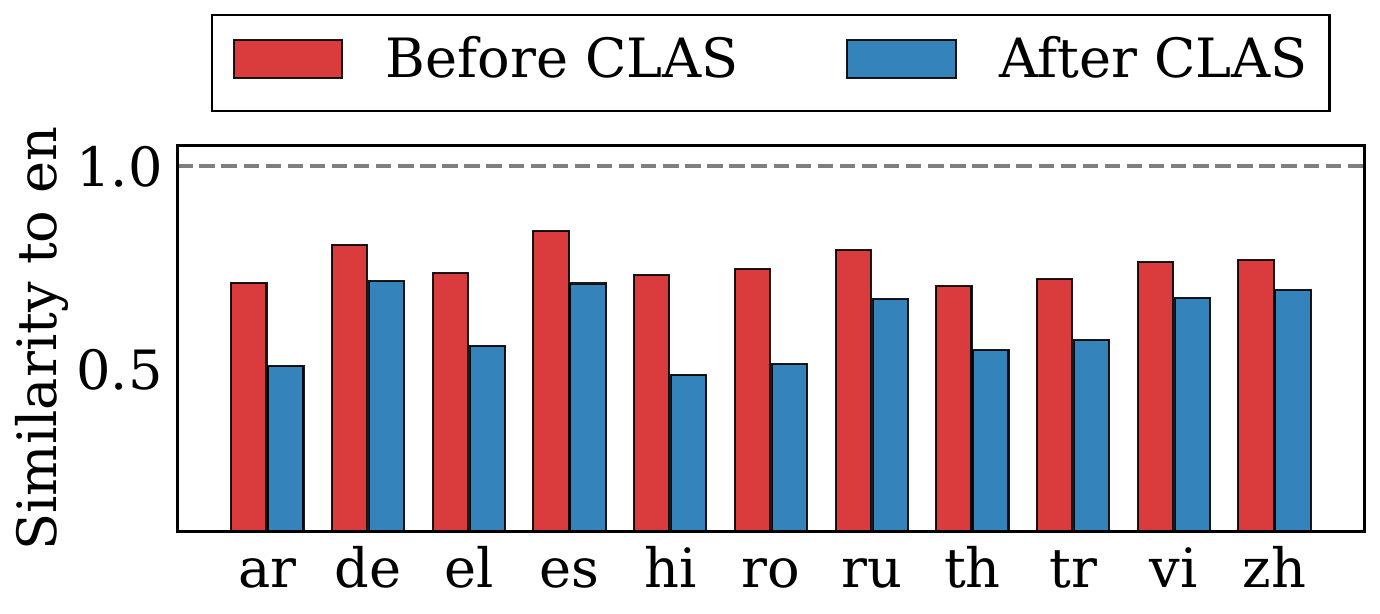}
        \caption{XQuAD}
        \label{fig:analysis_xquad_llama}
    \end{subfigure}

    \caption{Cosine similarity with English across languages on each task using Llama model.}\vspace{-0.5cm}
    \label{fig:cosine_sim}
\end{figure}

\begin{figure*}[!t]
    \centering
    
    \begin{subfigure}{0.48\textwidth} 
        \centering
        \includegraphics[width=0.49\linewidth]{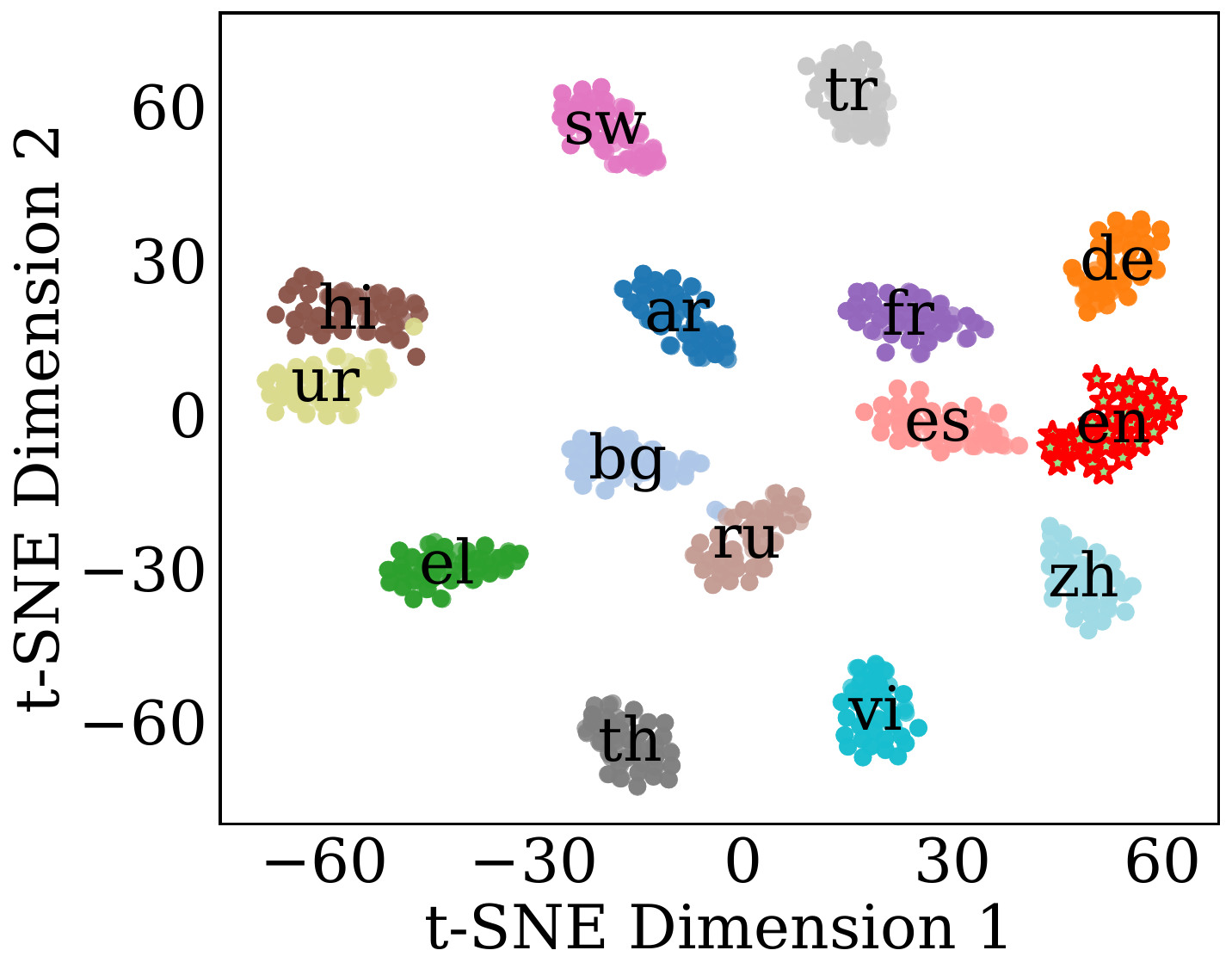}
        \hfill
        \includegraphics[width=0.49\linewidth]{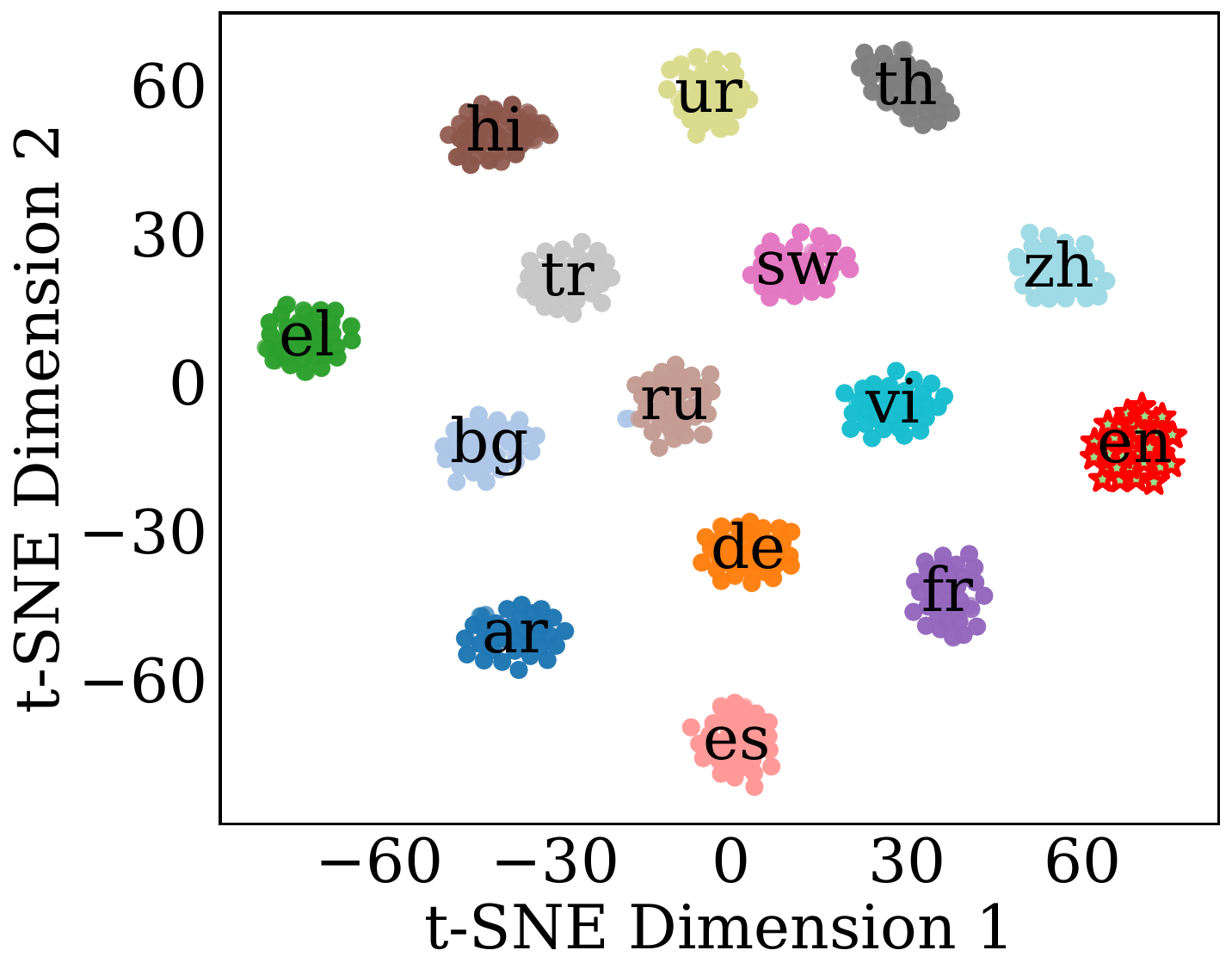}
        \caption{XNLI Llama (Before vs After)}
        \label{fig:xnli_llama_pair}
    \end{subfigure}\hfill
    \begin{subfigure}{0.48\textwidth} 
        \centering
        \includegraphics[width=0.49\linewidth]{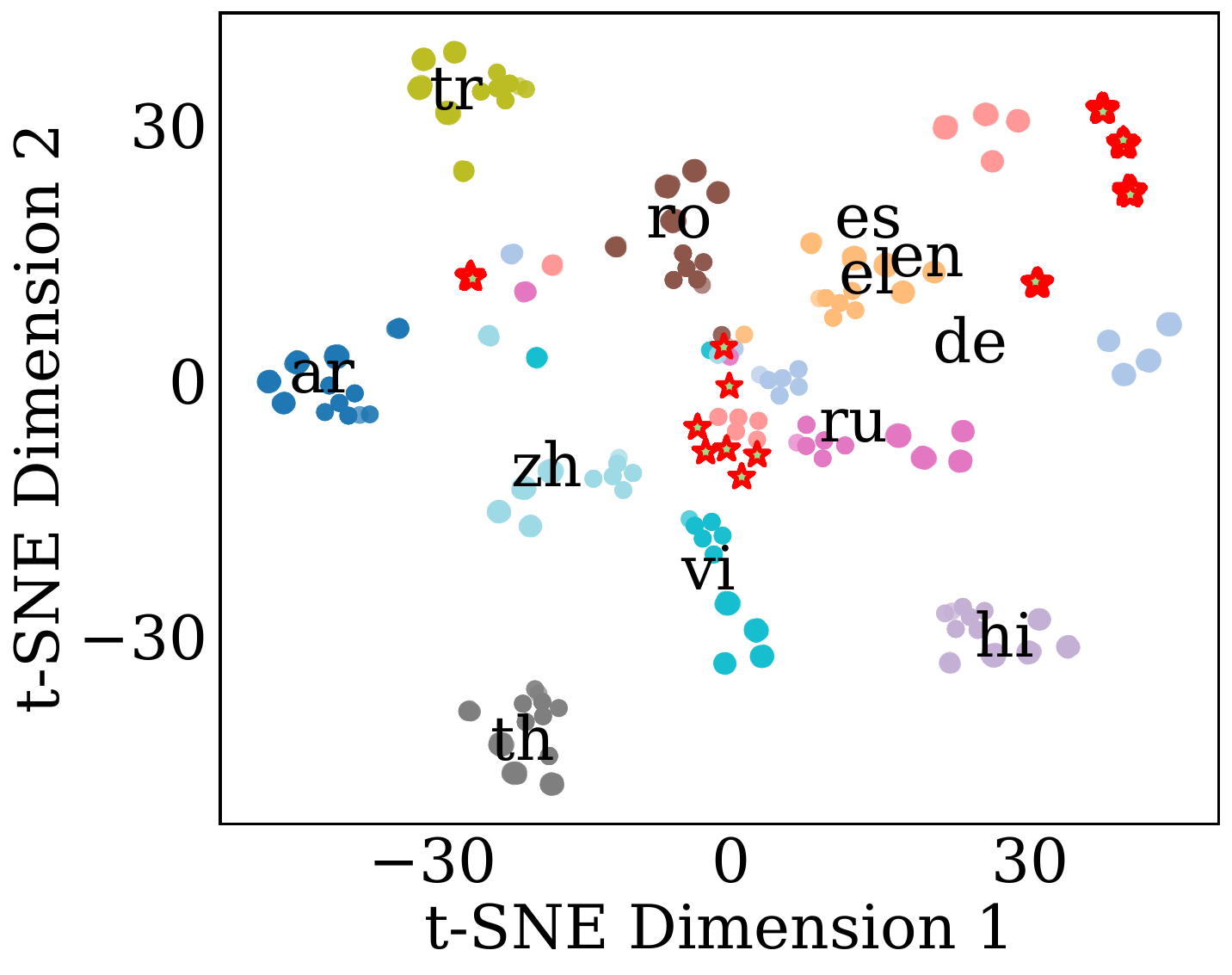}
        \hfill
        \includegraphics[width=0.49\linewidth]{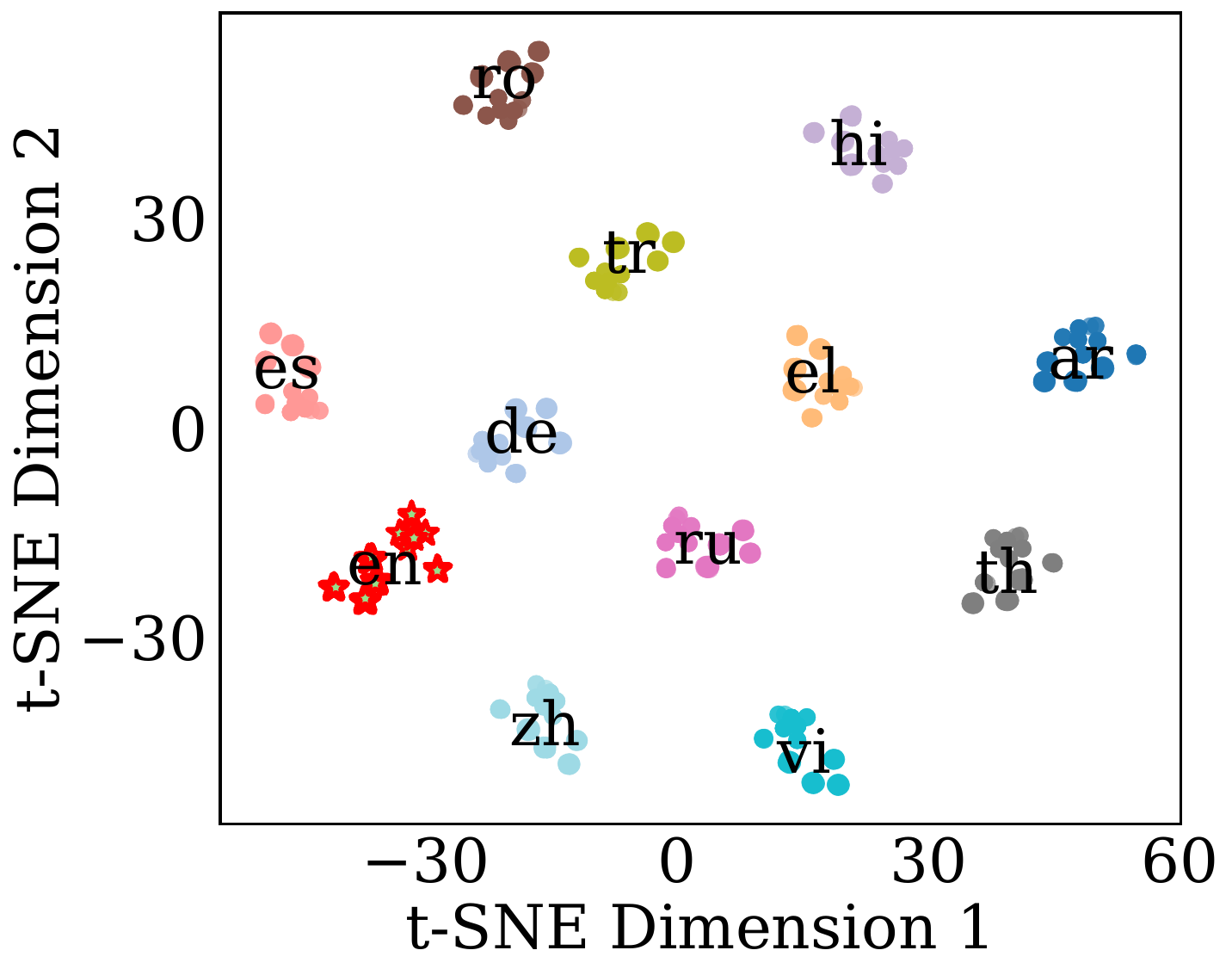}
        \caption{XQuAD Llama (Before vs After)}
        \label{fig:xquad_llama_pair}
    \end{subfigure}

    \caption{t-SNE visualization of multilingual representations before (left) and after (right) CLAS intervention. Each point represents a sentence embedding, with colors denoting languages. English, marked with (\textcolor{red}{red}) stars, is the anchor language.} \vspace{-0.4cm}
    \label{fig:tsne_before_after_xnli_xquad}
\end{figure*}


    


\subsection{Cross-Lingual Alignment Analysis}
In this subsection, we analyze the impact of our activation steering mechanism on representation alignment and examine how these geometric shifts relate to cross-lingual performance gains {on XNLI and XQuAD}.
Figure~\ref{fig:cosine_sim} shows cosine similarity between each target language and English in the bridge layers before and after CLAS. Across tasks and models, CLAS generally reduces similarity to English rather than increasing it. This effect is strongest for Llama, especially on XQuAD, where several languages move noticeably farther from English, while Qwen shows smaller and more uniform changes.

This pattern matches our neuron analysis presented earlier in Section~\ref{sec:neuron}. CLAS rebalances partial-shared and language-specific neurons, reducing over-reliance on English-centric features and allowing more target-language structure to influence predictions. As a result, CLAS improves performance not by pulling languages closer to English, but by relaxing excessive alignment. The effect is larger on XQuAD than on XNLI, likely because span extraction is more sensitive to surface form. Reducing English alignment can help match target-language expressions, though it can also increase variance and occasional regressions.

\begin{figure}[t!]
    \centering
    \vspace{-0.5em}
    \begin{subfigure}{0.2\textwidth}
        \centering
        \includegraphics[width=\linewidth]{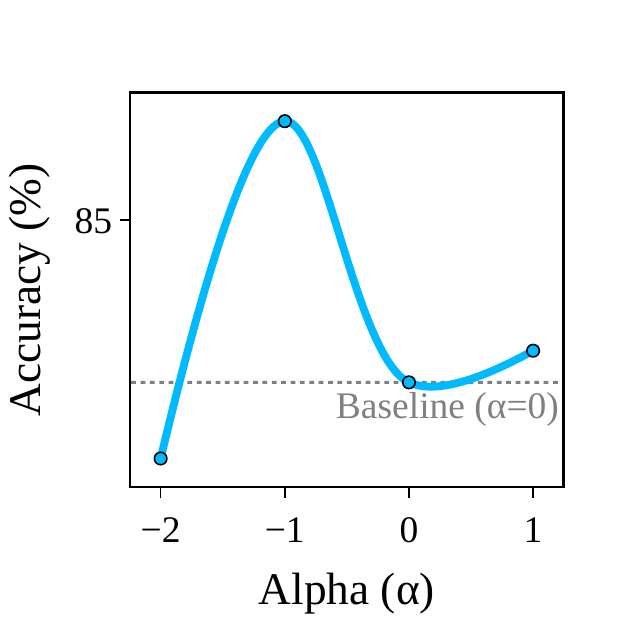}
        \caption{XStoryCloze (Llama)}
    \end{subfigure}\hspace{1em} 
    \begin{subfigure}{0.2\textwidth}
        \centering
        \includegraphics[width=\linewidth]{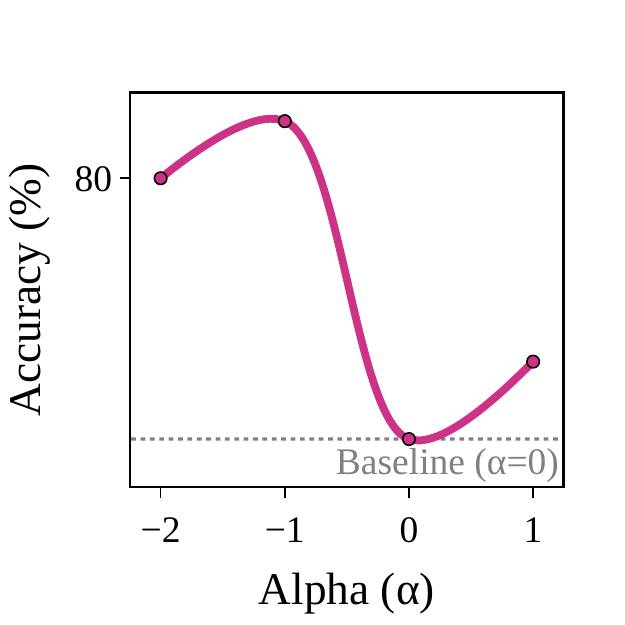}
        \caption{XStoryCloze (Qwen)}
    \end{subfigure}
    
    \vspace{-0.2em}
    
    \begin{subfigure}{0.2\textwidth}
        \centering
        \includegraphics[width=\linewidth]{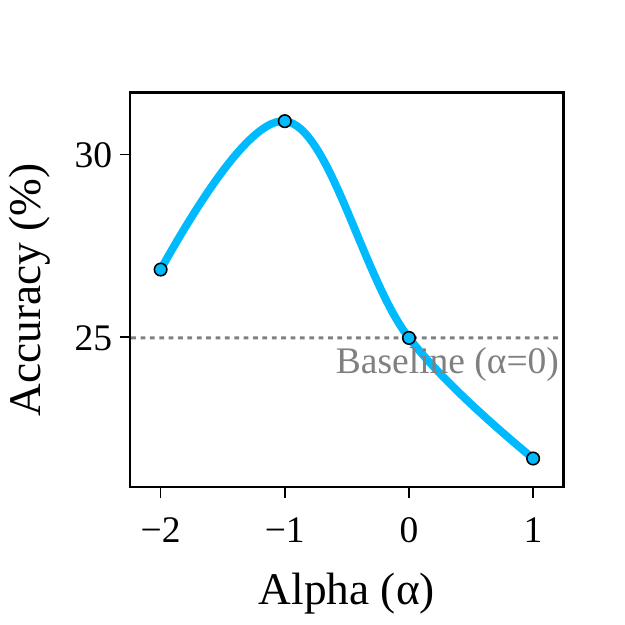}
        \caption{XQuAD (Llama)}
    \end{subfigure}\hspace{1em} 
    \begin{subfigure}{0.2\textwidth}
        \centering
        \includegraphics[width=\linewidth]{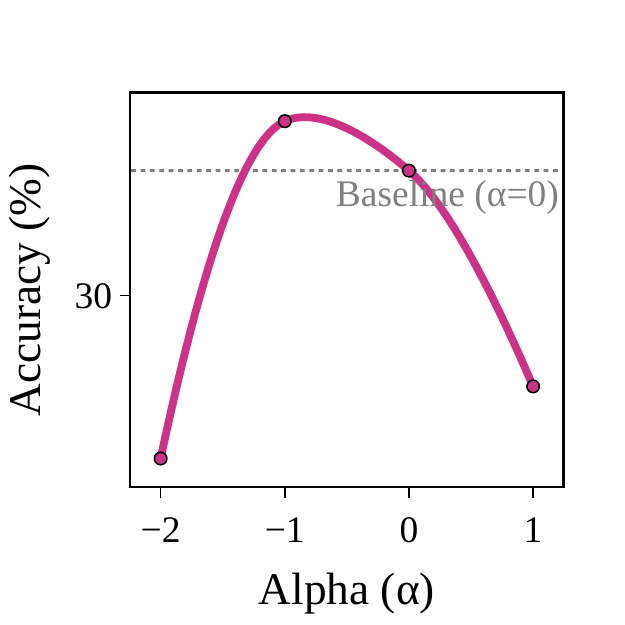}
        \caption{XQuAD (Qwen)}
    \end{subfigure}
    
    \vspace{-0.2em}
    
    \caption{Impact of $\alpha$ across XStoryCloze and XQuAD. The left column shows the performance for the Llama (\textcolor[HTML]{00BAFF}{blue}) and the right column for the Qwen (\textcolor[HTML]{CC3386}{pink}).} \vspace{-0.8cm}
    \label{fig:alpha_xstory_xquad}
\end{figure}

\subsection{Representation Space Visualization}

We use t-SNE to project the penultimate layer's hidden states for the English anchor and all other languages into a shared 2D space. Figure~\ref{fig:tsne_before_after_xnli_xquad} shows the representations and centroids before and after CLAS, revealing three patterns.


\paragraph{CLAS reorganizes representations without collapsing them into English.} 
After CLAS, language clusters remain distinct and are often more clearly separated, suggesting gains come from restructuring language-specific spaces rather than forcing assimilation to English.

\paragraph{Moderate reorganization correlates with larger gains.} 
On XNLI, increased cluster structure is broadly helpful. On XQuAD, the largest gains occur when CLAS resolves overlap for languages that start in mixed regions (e.g., \textit{ar}, \textit{de}, \textit{es}); already well-separated languages (e.g., \textit{hi}, \textit{vi}, \textit{zh}) change less and typically gain less. This suggests that CLAS is most helpful when it resolves representational overlap, rather than when representations are already well-formed.


\paragraph{Distance to English is not predictive.}
Neither initial nor final proximity to the English cluster reliably explains performance changes. Some languages far from English (e.g., \textit{ur}, \textit{hi}) improve largely as seen in \citep{lim2025language}, while some closer languages improve little or regress. After CLAS, high-performing languages are scattered rather than concentrated near English. This suggests that CLAS effectiveness depends on how representations are reorganized, not simply on how close they are to the anchor language.

\begin{table*}[!t]
\centering
\small
\begin{tabular}{p{4.8cm}p{6.8cm}p{3cm}}
\toprule
\textbf{Question} &\textbf{Llama (base)} &\textbf{CLAS} \\
\midrule
Was benannte die Generalkonferenz für Maß und Gewicht 1960 nach Tesla? &SI-Maßeinheit der magnetischen Flussdichte. stop. Die SI-Maßeinheit der magnetischen Flussdichte &SI-Maßeinheit der magnetischen Flussdichte. \\
\textcolor{gray}{(What did the General Conference on Weights and Measures name after Tesla in 1960?)} &\textcolor{gray}{(SI unit of magnetic flux density. stop. The SI unit of magnetic flux density)} &\textcolor{gray}{(SI unit of magnetic flux density.)} \\
\midrule
Die Lutherbibel beeinflusste wessen englische Übersetzung der Bibel? &William Tyndale Die Lutherbibel beeinflusste die englische Bibel von William Tyndale. Tyndale war ein eng &William Tyndale. \\
\textcolor{gray}{(The Luther Bible influenced whose English translation of the Bible?)} &\textcolor{gray}{(William Tyndale's English Bible was influenced by the Luther Bible. Tyndale was a close)} &\textcolor{gray}{(William Tyndale.)} \\
\bottomrule
\end{tabular}
\caption{Comparison of outputs for German XQuAD samples. English translations provided via Google Translate.}\vspace{-0.4cm}
\label{tab:qualitative_german}
\end{table*}

\subsection{Optimal Steering Intensity and Direction}
\label{subsec:optimal_alpha}

As shown in Figure~\ref{fig:alpha_xstory_xquad}, Llama and Qwen both achieve peak accuracy on XStoryCloze and XQuAD when $\alpha = -1$. Under the CLAS formulation, negative $\alpha$ to suppresses language-specific neurons. Since $-\lambda$ already regulates dominant shared neurons, this trend indicates that these tasks benefit most from masking language-specific signals and relying entirely on shared semantic structures. However, optimal steering varies. Appendix~\ref{sec:alpha_search} shows the settings that benefit from positive steering, where selectively amplifying language-specific neurons captures necessary target-language nuances. This aligns with prior findings that optimal cross-lingual transfer remains inherently task-dependent \cite{tang-etal-2024-language, wang2024sharing}.

\subsection{Importance of Bridge Layer Selection}
\begin{figure}[!t]
    \centering
    \includegraphics[width=0.4\textwidth]{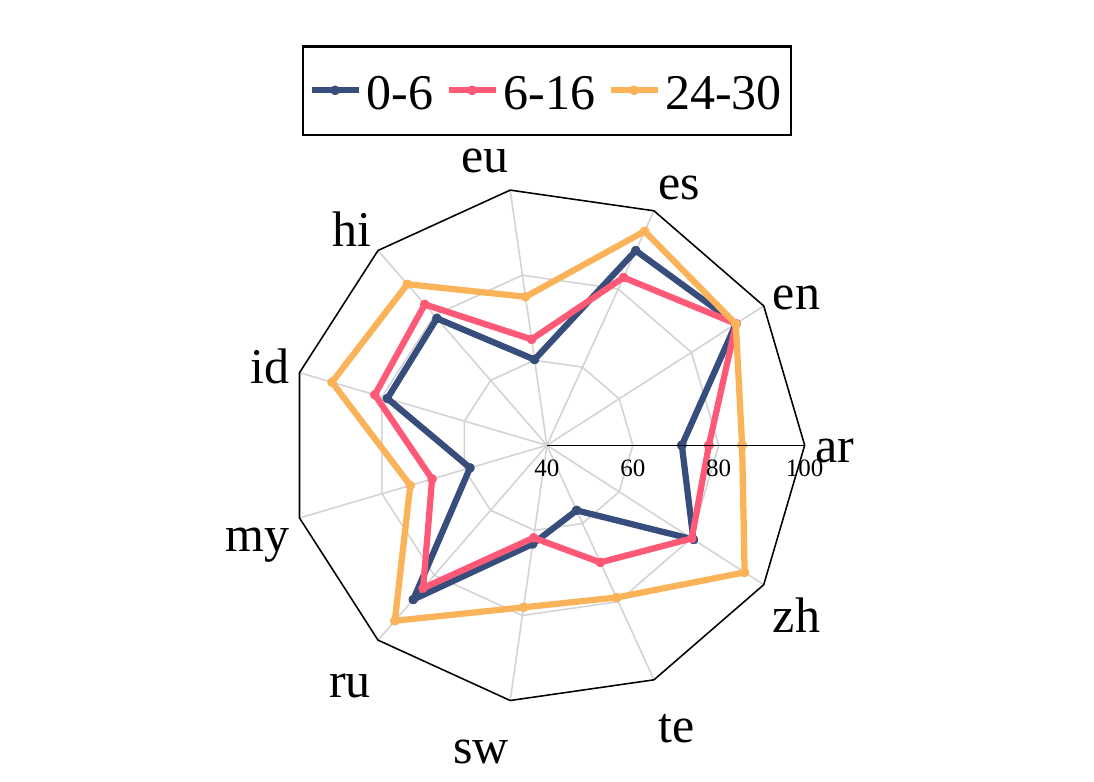}
    \caption{Multilingual performance comparison across different Llama layers (early layers 0-6, middle layers 6-16, and bottom layers 24-30).}
    \label{fig:bridge_layer_xstory}\vspace{-0.4cm}
\end{figure}

To study whether intervening on other layers produces comparable results, we evaluate the impact of CLAS on the XStoryCloze with Llama. Specifically, we compare our selected bridge layers (24–30) against two distinct representational regions: the early layers (0–6), which contain a higher density of language-specific neurons, and the middle layers (6–16), where all-shared neuron activity is the highest. As illustrated in Figure~\ref{fig:bridge_layer_xstory}, applying the intervention to the early or middle layers yields heavily suboptimal cross-lingual performance, achieving average scores of only 72.57 and 75.43, respectively. Intervening too early likely disrupts low-level lexical processing \cite{tang-etal-2024-language,kojima-etal-2024-multilingual}, while suppressing the shared neurons in the middle layers risks degrading the model's language-agnostic representations \cite{xu-etal-2025-linguistic, zhang2025does}. In contrast, applying CLAS to the bridge layers (24–30) results in a performance spike, reaching an average of 85.58.  This demonstrates that identifying the correct intervention site is critical; the bridge layers provide the optimal semantic space where shared representations can be safely dampened and allow language-specific features to steer the final output distribution.

\subsection{Qualitative Analysis}

Table \ref{tab:qualitative_german} shows the qualitative impact of CLAS on German generation. In the examples, the baseline model suffers from severe repetition loops and verbosity leakage. In contrast, CLAS successfully suppresses these behaviors. This demonstrates CLAS effectively generates concise response while maintaining accuracy.



\section{Related Work}
\paragraph{Cross-lingual Transfer}
Cross-lingual transfer in multilingual LLMs has been widely studied, with prior work showing that transfer quality varies across tasks, languages, and models. For example, \citet{hu-etal-2025-large-language} analyze factors that influence cross-lingual performance on reasoning tasks. Many approaches improve transfer through additional training such as adding more data during instruction tuning \citep{shaham-etal-2024-multilingual}, combining supervised fine-tuning with preference alignment \citep{lai-etal-2024-llms}, and constructing new multilingual pre-training datasets \citep{he2025semantic}. Other work uses translation-based fine-tuning \citep{lee-etal-2025-cross}, layer-wise fine-tuning \citep{bandarkar2025layer}, or language-specific adapters \citep{zhao-etal-2025-adamergex}. Continued pre-training has also been shown to improve transfer for some language pairs \citep{wu-etal-2025-enhancing-llm}. An alternative line of work uses prompts rather than parameter updates. For example, \citet{tanwar-etal-2023-multilingual} use multilingual in-context examples, and \citet{yoo2025code} study in-context learning in code-switching settings.

\paragraph{Neuron Behavior and Cross-lingual Transfer}
Several studies examine cross-lingual transfer at the neuron level. \citet{huang-etal-2025-neurons} show that activation similarity across languages is associated with better transfer. Other work identifies language-specific and shared neurons \citep{tang-etal-2024-language, zhang2025does, tezuka-inoue-2025-transfer}, and shows that shared neurons often concentrate in middle and upper layers \citep{xu-etal-2025-linguistic}. Results on directly intervening on neurons are mixed. \citet{wang2024sharing} show that neuron roles vary by task and model while \citet{mondal-etal-2025-language} find that manipulating language-specific neurons yields limited gains. Our work differs in that we apply test-time steering that blends rather than overwrites activations, preserving representational structure. We show that the direction and magnitude of steering matter, and that carefully controlled neuron-level interventions can improve cross-lingual transfer. 
\citet{sundar-etal-2025-steering} target language-specific neurons to push multilingual embedding into a shared aligned space whereas we show that effective cross-lingual transfer actually relies on maintaining distinct separation between language clusters.
Lastly, INCLINE \cite{wang-etal-2025-bridging} learns per-layer alignment matrices at inference time but requires per-language tuning. On the other hand, CLAS tunes a single parameter per task.

\section{Conclusion}
We presented CLAS, a training-free activation steering method that improves cross-lingual transfer by rebalancing shared and language-specific neurons at inference time. CLAS improves performance across both discriminative and generative tasks, and our analysis shows that gains come from functional divergence rather than forcing representations to align closely with the anchor language. 
Languages benefit most by shifting from ambiguous regions into distinct clusters, whereas initial anchor proximity does not reliably predict success.
These patterns are task-dependent, highlighting the need for task-aware cross-lingual methods. Future work could explore adaptive steering strategies that adjust intervention strength based on representational structure, better understand saturation effects in already well-separated clusters, and extend CLAS to other modalities and training settings.

\section*{Limitations}
{We evaluate CLAS on two multilingual instruction-tuned LLMs (Qwen and Llama) and across multiple multilingual benchmarks. However, our analysis and suggested intervention are anchored to English i.e. we quantify alignment shifts via cosine similarity to an English anchor. This may not reflect behavior under alternative references/anchors or truly language pair-specific settings. Additionally, it is important to note that CLAS is not uniformly beneficial: while average gains for languages are positive, we also observe language-specific regressions (e.g, Hindi, Greek). This indicates that test-time steering can be brittle for certain model/language combinations. Finally, our mechanistic analysis does not isolate the role of attention heads or other circuit components, limiting the granularity of causal attribution.}

\section*{Ethical Considerations}
{CLAS is a test-time activation intervention and can change model behavior in ways that are not always predictable across languages, including occasional performance degradations. As with other steering approaches, such interventions can be re-purposed in undesirable ways (e.g, modulating output without transparency). Thus, we recommend caution and task-specific testing and validation before real-world deployment. Our experiments use publicly-available benchmarks and do not involve human subjects or personal data.}

\section*{Acknowledgments}
We appreciate the valuable feedback from the anonymous reviewers and the members of the PortNLP Lab. We also gratefully acknowledge the support of the David E. Wedge Awards for Excellence from Portland State University’s Maseeh College of Engineering and Computer Science.

\bibliography{custom}

\appendix

\section{Dataset Summary}\label{app:dataset}
Table~\ref{tab:dataset_summary} presents an overview of the datasets used in this study.

\begin{table}[!t]
\centering
\setlength{\tabcolsep}{0.25pt}
\small
\begin{tabular}{lccc}
\toprule
\textbf{Dataset} & \textbf{$|\mathcal{L}|$} & \textbf{\#Samp.} & \textbf{Metric}\\
\midrule
XNLI   (natural language inference)     & 15 & 5010 & Acc.\\
XStoryCloze (commonsense reasoning) & 11 & 1511 & Acc.\\
PAWS-X  (paraphrase identification)    & 4  & 2000 & Acc. \\
HarryPotter (MCQA) & 5  & 300  & Acc.\\
XQuAD (question answering)      & 12 & 1190 & F1\\
\bottomrule
\end{tabular}
\caption{Summary of evaluation datasets. \#Samp. represents the number of samples per language.} 
\label{tab:dataset_summary}
\end{table}

\section{Language-Level Performances}
\label{sec:rem_lang_peroformance}
Tables~\ref{tab:xnli_results}, \ref{tab:pawsx_results}, and \ref{tab:harrypotter_results} show the language level results on XNLI, PAWS-X, and HarryPotter Quiz across Llama and Qwen models.

\section{Effect of $\alpha$}
\label{sec:alpha_search}
\begin{figure*}[t!]
    \centering
    
    \begin{subfigure}{0.25\textwidth}
        \centering
        \includegraphics[width=\linewidth]{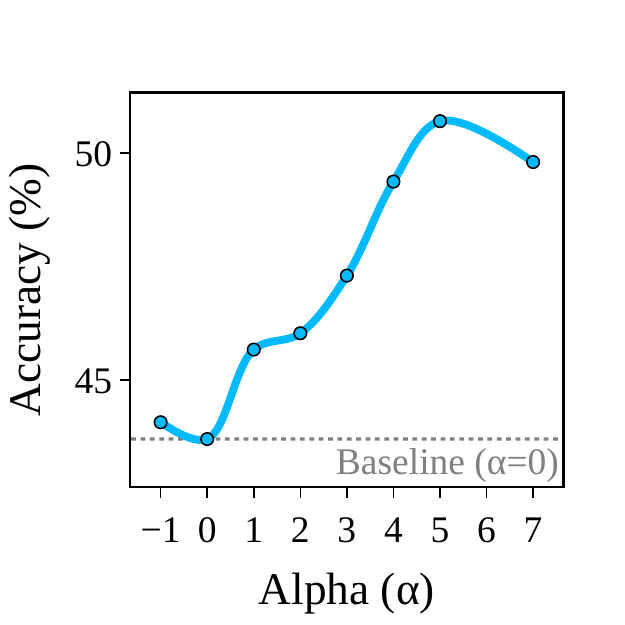}
        \caption{XNLI (Llama)}
    \end{subfigure}\hspace{1em}
    \begin{subfigure}{0.25\textwidth}
        \centering
        \includegraphics[width=\linewidth]{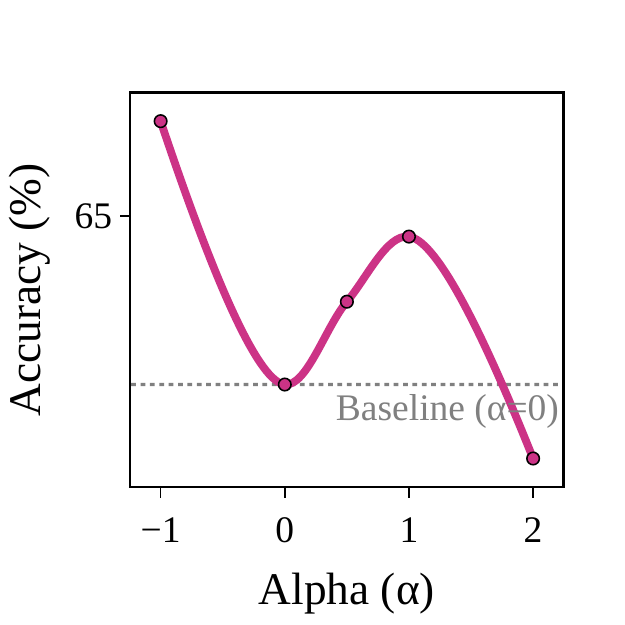}
        \caption{XNLI (Qwen)}
    \end{subfigure}
    
    
    \begin{subfigure}{0.25\textwidth}
        \centering
        \includegraphics[width=\linewidth]{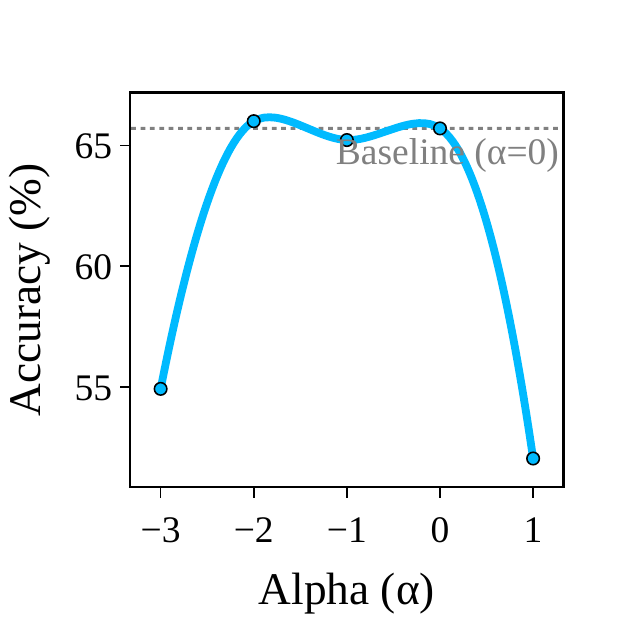}
        \caption{PAWS-X (Llama)}
    \end{subfigure}\hspace{1em}
    \begin{subfigure}{0.25\textwidth}
        \centering
        \includegraphics[width=\linewidth]{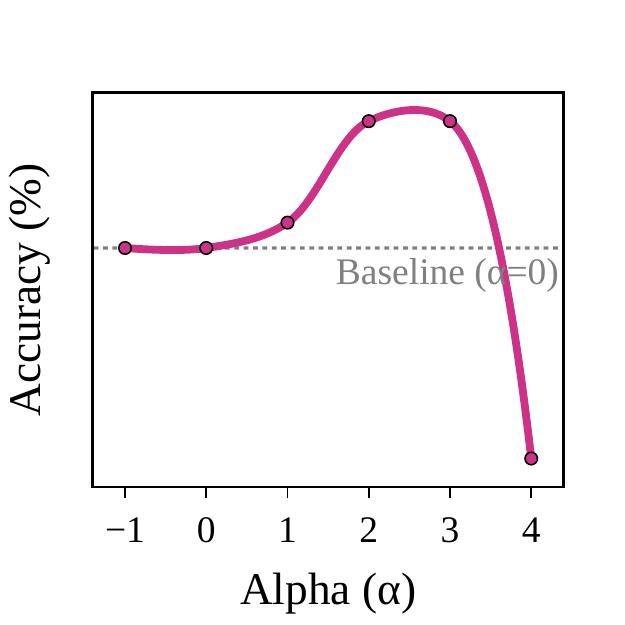}
        \caption{PAWS-X (Qwen)}
    \end{subfigure}
    

    \begin{subfigure}{0.25\textwidth}
        \centering
        \includegraphics[width=\linewidth]{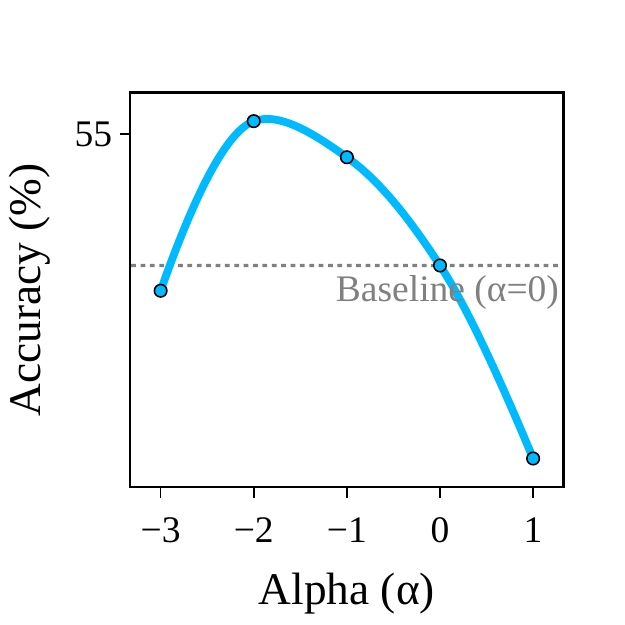}
        \caption{HarryPotter Quiz (Llama)}
    \end{subfigure}\hspace{1em}
    \begin{subfigure}{0.25\textwidth}
        \centering
        \includegraphics[width=\linewidth]{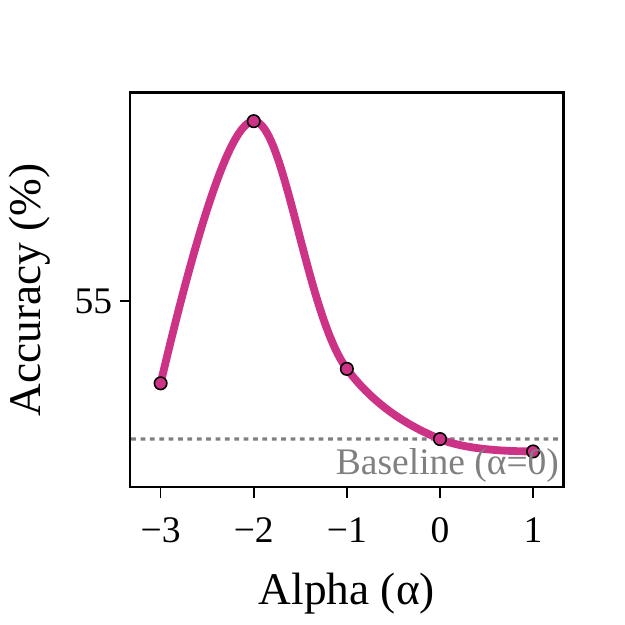}
        \caption{HarryPotter Quiz (Qwen)}
    \end{subfigure}
    
    \vspace{1em}

    \caption{Impact of the hyperparameter $\alpha$ on performance across datasets. The left column shows the performance for the Llama model, while the right column shows the performance for the Qwen model.}
    \label{fig:alpha_all_datasets}
\end{figure*}

Figure~\ref{fig:alpha_all_datasets} shows the effect of $\alpha$ on XNLI, PAWS-X and HarryPotter Quiz datasets.




\begin{table*}[!t]
\centering
\small
\begin{tabular}{p{0.9cm} p{0.5cm} p{0.5cm} p{0.5cm} p{0.5cm} p{0.5cm} p{0.5cm} p{0.5cm} p{0.5cm} p{0.5cm} p{0.5cm} p{0.5cm} p{0.5cm} p{0.5cm} p{0.5cm} p{0.5cm} p{0.5cm}}
\toprule

&en &ar &bg &de &el &es &fr &hi &ru &sw &th &tr &ur &vi &zh &\textbf{Avg.} \\
\midrule
\multicolumn{17}{l}{\textit{Llama}} \\
\cmidrule(lr){1-17}
Llama &52.63 &38.88 &42.28 &43.65 &40.58 &46.95 &44.53 &41.78 &43.75 &41.10 &40.68 &44.17 &36.09 &44.81 &44.63 &42.42 \\
Int $\mu$ &52.63 &38.12 &40.68 &45.77 &41.3 &46.59 &43.11 &41.10 &43.79 &41.42 &44.39 &43.81 &36.37 &42.85 &41.38 &42.19 \\
INCLINE &52.63 &39.24 &38.74 &35.77 &38.08 &38.62 &35.75 &44.25 &33.51 &35.51 &42.87 &39.60 &33.35 &46.89 &46.43 &39.19 \\
CLAS &52.63 &39.10 &41.68 &49.42 &44.85 &50.38 &47.96 &41.74 &45.27 &38.36 &40.98 &49.36 &44.25 &46.41 &50.72 &\textbf{45.03} \\
\addlinespace
\multicolumn{17}{l}{\textit{Qwen}} \\
\cmidrule(lr){1-17}
Qwen &71.74 &60.45 &60.75 &66.22 &58.02 &65.56 &64.47 &55.56 &61.87 &40.31 &57.8 &59.18 &51.07 &62.11 &63.81 &59.08 \\
Int $\mu$ &71.78 &60.54 &60.74 &66.33 &58.06 &65.51 &64.33 &55.51 &61.70 &39.96 &57.78 &59.10 &51.16 &62.18 &63.81 &59.05 \\
INCLINE &71.74 &54.51 &44.71 &58.14 &33.37 &33.33 &56.59 &33.33 &60.82 &34.07 &33.33 &33.33 &33.33 &36.55 &33.33 &41.34 \\
CLAS &71.74 &60.64 &60.94 &66.87 &57.92 &65.99 &65.21 &56.21 &61.86 &39.90 &57.09 &59.40 &50.98 &61.68 &65.35 &\textbf{59.29} \\
\bottomrule
\end{tabular}
\caption{The performance on XNLI across multiple languages and models.}
\label{tab:xnli_results}
\end{table*}

\begin{table*}[!t]
\centering
\small
\begin{tabular}{lrrrrr}
\toprule
 & en & de & es & fr & \textbf{Avg.} \\
\midrule
\multicolumn{6}{l}{\textit{Llama}} \\
\cmidrule(lr){1-6}
Base   &68.20 &63.45 &64.45 &69.20 &65.70 \\
Int $\mu$  &68.20 &62.20 &54.70 &71.70 &62.87 \\
INCLINE &69.00 &66.45 &54.55 &57.75 &59.58 \\
CLAS    &68.20 &66.60 &65.70 &65.70 &\textbf{66.00} \\
\addlinespace
\multicolumn{6}{l}{\textit{Qwen}} \\
\cmidrule(lr){1-6}
Base    &84.20 &78.85 &78.75 &78.15 &78.58 \\
Int $\mu$  &84.20 &79.00 &79.85 &78.25 &79.03 \\
INCLINE &84.50 &79.95 &78.70 &46.25 &68.30 \\
CLAS    &84.20 &79.55 &79.25 &78.40 &\textbf{79.07} \\
\bottomrule
\end{tabular}
\caption{Performance on the PAWS-X across multiple languages and models.}
\label{tab:pawsx_results}
\end{table*}

\begin{table*}[t]
\centering
\small
\begin{tabular}{lrrrrrr}
\toprule
 & en & de & fr & it & es & \textbf{Avg.} \\
\midrule
\multicolumn{7}{l}{\textit{Llama}} \\
\cmidrule(lr){1-7}
Base   &65.00 &50.67 &50.67 &48.33 &56.67 &51.59 \\
Int $\mu$  &65.33 &53.67 &44.33 &46.67 &46.67 &47.84 \\
INCLINE &61.00 &52.00 &49.00 &48.67 &53.33 &50.75 \\
CLAS    &65.00 &53.67 &52.00 &47.67 &57.00 &\textbf{52.59} \\
\addlinespace
\multicolumn{7}{l}{\textit{Qwen}} \\
\cmidrule(lr){1-7}
Base    &57.00 &55.00 &51.33 &48.33 &60.00 &53.67 \\
Int $\mu$  &53.67 &51.33 &50.67 &41.00 &55.33 &49.58 \\
INCLINE &58.67 &24.33 &40.67 &24.33 &24.33 &28.42 \\
CLAS    &57.00 &59.00 &54.33 &47.67 &61.00 &\textbf{55.50} \\
\bottomrule
\end{tabular}
\caption{Accuracy on the multilingual HarryPotter Quiz across multiple languages and models.}
\label{tab:harrypotter_results}
\end{table*}

\end{document}